\documentclass{article}

\usepackage{PRIMEarxiv}

\usepackage[utf8]{inputenc} 
\usepackage[T1]{fontenc}    
\usepackage{hyperref}       
\usepackage{float}
\usepackage{mathtools}
\usepackage{url}       
\usepackage{booktabs}       
\usepackage{amsfonts}       
\usepackage{bbm}
\usepackage{subcaption}
\usepackage{nicefrac}       
\usepackage{microtype}      
\usepackage{lipsum}
\usepackage{mathrsfs}
\usepackage{algorithm2e}
\usepackage{algcompatible}
\RestyleAlgo{ruled} 
\usepackage{fancyhdr}       
\usepackage{graphicx}       
\graphicspath{{media/}}     

\title{ A computational exploration of emerging methods of variable importance estimation
}

\author{
 Louis Mozart Kamdem \\
  AIMS-Rwanda\\
  Kigali\\
  \texttt{louismozart.teyou@aims.ac.rw} \\
   \And
  Ernest Fokoue \\
 Rochester University of Technology (RIT) \\
  New-York\\
  \texttt{epfaqa@rit.edu} \\
}

\begin{document}
\maketitle

\begin{abstract}
Estimating the importance of variables is an essential task in modern machine learning. This help to evaluate the goodness of a feature in a given model. Several techniques for estimating the importance of variables have been developed during the last decade. In this paper, we proposed a computational and theoretical exploration of the emerging methods of variable importance estimation, namely: Least Absolute Shrinkage and Selection Operator (LASSO), Support Vector Machine (SVM), the Predictive Error Function (PERF), Random Forest (RF), and Extreme Gradient Boosting (XGBOOST) that were tested on different kinds of real-life and simulated data. All these methods can handle both regression and classification tasks seamlessly but all fail when it comes to dealing with data containing missing values. The implementation has shown that PERF has the best performance in the case of highly correlated data closely followed by RF. PERF and XGBOOST are "data-hungry" methods, they had the worst performance on small data sizes but they are the fastest when it comes to the execution time. SVM is the most appropriate when many redundant features are in the dataset. A surplus with the PERF is its natural cut-off at zero helping to separate positive and negative scores with all positive scores indicating essential and significant features while the negatives score indicates useless features. RF and LASSO are very versatile in a way that they can be used in almost all situations despite they are not giving the best results.\end{abstract}


\section{Introduction}

The modern world is fully under the influence of artificial intelligence, which consists of systems or machines to imitate human intelligence. These systems learn and improve their performance at the same time according to the data they process via statistical models. With the emergence of new information and communication technologies such as smartphones, powerful computers, and satellite devices, data collection is becoming increasingly easy, which sometimes results in very large-scale data that can range from tens to hundreds of thousands of variables. However, in the context of linear models, Generalized linear models, nonparametric and parametric models, it is sometimes important and even crucial to measure the impact of variables in those models. This process is known as variable ranking. The variable ranking is the process of ordering features based on the value of a scoring function that measures feature relevance. Most of the time, variable ranking is followed by variable selection. In reality, after arranging the variables in order of importance, it is wise to eliminate those variables that do not contribute much to a good performance of the model used (fewer importance variables). This will reduce the number of variables and therefore improve the performance of the model. The selection of variables is very important in many scientific fields such as bioinformatics. Variable and feature selection have become the focus of many subfields of bioinformatic such as Gene expression classification, Sequence annotation, and Statistical genetic.

\textbf{Gene expression classification} 

The advent of DNA technology \cite{saeys2007review} made researchers able to measure the expression levels of many genes at the same time. Most microarray gene expression datasets suffer from the well-known problem of the "small $n$ large $p$". That is, the dimensions are very large (sometimes in the range of thousands) and the sample size is small (usually up to hundreds).
An important task in biomedical research is to classify various types of disease \cite{statnikov2008comprehensive}. In order to use gene expression data for the classification of diseases against nondisease samples, \cite{lee2005extensive} conducted extensive research to compare the KNN approach, different versions of Linear Discriminant Analysis (LDA), bagging trees, boosting, and RF in the same experimental setup. They found that RF was the most successful tool to analyse the seven microarray datasets. 
Due to the high dimensional and small size of Gene expression data sets, fast and efficient variable importance measures techniques have received a great deal of attention. Genetic interactions are an important factor to consider when measuring feature scores. However, famous univariate variable measure strategies like variance-based variable importance couldn't be taken into consideration. Therefore, researchers have proposed other techniques for capturing correlations between genes using  variable importance estimations like random forest variable importance (\cite{diaz2005variable}, \cite{amaratunga2008enriched}). For example \cite{amaratunga2008enriched} proposed the "enriched random forest" method. This has improved the RF performance on the ten real gene expression data sets since it uses a random sampling scheme to select the features with the highest score. The use of microarrays for gene expression has emerged as a popular tool for identifying common genetic factors that affect our health and disease. The variable importance measurement can provide good issues for analyzing and identifying a patient's molecular profile from gene expression data sets \cite{saeys2007review}.

\textbf{Sequence annotation} 

Biological sequences analysis is still a growing area and emerges sub-field in bioinformatics. Biological mutations and evolution made too many species genomes to be sequenced \cite{qi2012random}. Besides the basic features such as nucleotide or amino acids at each position in a sequence, many other features can be derived and their numbers grow exponentially \cite{saeys2007review}. Since many of them are useless or redundant, different variables' important measures  techniques are applied among others to focus only on the set of important variables.

\textbf{Statistical genetics}  

Statistical biology is a type of computational biology where different statistical methods are applied to draw inferences from genetic data \cite{saeys2007review}. In the realms of healthcare a genetic data is unquestionably a valuable source of information. Governments, on the other hand, are increasingly looking to develop and explore massive databases of genetic profiles for a variety of reasons. For instance, genetic data is often used for a criminal investigation to prove the culpability of a person, and this is without his contentment. The complexity and a huge number of sequences in such kinds of data require feature extractions.

The general formulation of a data is the following:
\begin{equation}\label{1.1.1}
	\mathscr{D}_{n} = \biggl\{(\mathbf{x}_i,\mathbf{y}_i)\sim\mathbb{P},\mathbf{x}_i\in\mathscr{X},\mathbf{y}_i\in\mathscr{Y},i = 1,2,\dots,n\biggr\}
\end{equation}

\begin{itemize}
	\item all the $\mathbf{x}_i, i = 1,2,\dots,n $ are the input variables or simply features
	\item $ \mathscr{X}  $ is called the input space
	\item $\mathbf{y} = (y_1,\dots,y_n)^T$  is the response variable
     \item  $\mathbf{X} = (\mathbf{x}_1,\dots,\mathbf{x}_n)$ is the data matrix
	\item $ \mathscr{Y} $ is called the output space
	\item $\mathbb{P}$ is a probability distribution followed by the data to ensure randomness.
\end{itemize}

In the view of machine learning, it is interesting to understand the relationship between $\mathbf{y}$ and $\mathbf{X}$ in order to make prediction. For that, we consider the conditional mean function $f(\mathbf{x}) = E_\mathbb{P}(\mathbf{y}|X = \mathbf{x})$ that we need to estimate. This is known as the predictive modeling problem. Some techniques like linear regression, random forest, artificial neural network,  and so on can be used to estimate $f$. As we can guess, the estimation of $f$ entirely depends on the inputs variables which can be very large in many applications. Therefore, it is crucial to select variables that are most significant to the variation of $f$. Doing so will help to improve the performance of the built model by reducing the training time, making the model simpler, accurate and easy to interpret. To do so, many variables importance estimation techniques have been developed over the last decades. Among them, we have the Least Absolute Shrinkage and Selection Operator (LASSO), Support Vector Machine (SVM), the Predictive Error Function (PERF), Random Forest (RF), and Extreme Gradient Boosting (XGBOOST).

This work is motivated by a computational exploration and comparison of those techniques through different kinds of data such as High correlated data, Ultra-high dimensional data, Infra-high dimensional data, and data with many redundant features.

This paper is organized as follows. In section \ref{sec:headings}, we describe some previous work around the variable importance estimation, then in section \ref{sec1:headings}, we provide a mathematical background of some of the emerging methods use in the machine learning framework to estimate the importance of variables. In section \ref{sec2:headings} we showed the result of the implementation of those methods through different kind of data sets. Finally in section \ref{sec3:headings}, we provide a discussion and the conclusion of the work.

\section{Previous works}
\label{sec:headings}

\subsection{The Variance-based variable's important estimation}

The variance-based variable importance estimation is among the main and old classical tools for evaluating variable importance. \cite{mckay1997nonparametric} access the relative importance of an input variable by measuring the variance of the model output described by that variable. considering the data structure in \ref{1.1.1}, the measure of the importance of a variable $\textbf{x}$ with variance-based methods is given by:
\begin{equation}
\eta^2 = \frac{\mathbb{V}\big(E(\textbf{y}|\textbf{x})\big)}{\mathbb{V}(\textbf{y})}
\end{equation}
	
Under the assumption of the regression, \cite{mckay1997nonparametric} defines
\begin{equation}
   R^2 = \sum_{k=1}^{n}\hat{\beta}_k^2\mathbb{V}(\textbf{x}_k)/\hat{\mathbb{V}}(\textbf{y}) 
\end{equation}

where $\hat{\beta}_k$ is the estimate of the regressor ${\beta}_k$.
$ R^2 $ defined above is an estimator of $\eta^2$.

 Under the assumption that $E(\textbf{y}|\textbf{x})  =  \textbf{x}{\beta}$ (simple linear regression), we have: \begin{equation}
\rho^2 = \eta^2	
\end{equation}
where $\rho$ is the correlation coefficient define by:
\begin{equation}
	\rho = \frac{\sigma_{xy}}{\sigma_x\sigma_y}
\end{equation}


Therefore, for a simple linear regression model, the variance-based measure is simply the  correlation coefficient $\rho$. From the definition of $\rho$, it is obvious (using the Cauchy-Schwartz inequality) that $0\leq\rho\leq1$. For the values of $\rho$ close to one, the input variable is important to predict the output variable in a univariate linear regression and for the values of $\rho$ close to zero, the input variable is useless in the prediction of the output variable.

\subsection{Generalization of Variance-based variables important estimation}
Most variable important measures like  ANOVA-based variable importance measures are based on parametric assumptions  and this can be misleading. \cite{gromping2015variable} review in detail some of them in the case of a linear regression model.
Recent research has focused on expanding this definition by removing the parametric assumptions \cite{williamson2021nonparametric}. 

Using the same notation in \ref{1.1.1}, let consider the independent observations $\mathbf{Z}_1,\mathbf{Z}_2,\dots,\mathbf{Z}_n$ sampled from and unknown distribution $\mathbb{P}_0$. We define $\mathscr{M}$ as the class of potential distributions ($\mathscr{M}$ is our model). Each distribution $\mathbf{Z}_i$ consist of $(\mathbf{x}_i,\mathbf{y}_i)$ with $\mathbf{x}_i\in\mathscr{X}$ and $\mathbf{y}_i\in\mathscr{Y}$.

For all $\mathbb{P}\in\mathscr{M} $, we define $\mu_P(\mathbf{x}) = \mathbb{E}_\mathbb{P}(\mathbf{y}|X=\mathbf{x}) $. For a given set $s\subseteq\{1,\dots,p\}$ and $\mathbb{P}\in\mathscr{M}$ the reduced conditional mean is define by $\mu_{\mathbb{P},s}(\mathbf{x}) = \mathbb{E}_\mathbb{P}(\mathbf{y}|X_{-s}=\mathbf{x}_{-s}) $ . Given a set $r$ of indices and a vector $\mathbf{u}$, $\mathbf{u}_{-r}$ is the vector of all components of $u$ whose indices are not in $r$.

Example: Let $\mathbf{u} = (u_1,u_2,\dots,u_p)\in\mathbb{R}^p$ and $r = \{1,2,3\}$ then $\mathbf{u}_{-r} = (u_4,u_5,\dots,u_p)\in\mathbb{R}^{p-3} $  

Another formulation of the ANOVA-based variable important measure is:
\begin{equation}
	\psi_{0,s} = \frac{\int \{\mu_0(\mathbf{x})-\mu_{0,s}(\mathbf{x})\}^2d\mathbb{P}_{0}}{var_{\mathbb{P}_0}(\mathbf{y})} 
\end{equation}

where $ \mu_0(\mathbf{x}) = \mu_{\mathbb{P}_0}(\mathbf{x})$ and $\mu_{0,s}(\mathbf{x}) = \mu_{\mathbb{P}_{0,s}}(\mathbf{x})$. In the case of ANOVA, $u_0$ is assumed to have a simple parametric form which can be sometime misleading. reason why non parametric or machine learning  techniques are used to estimate $\mu_0$. Under the only restriction that for all $\mathbb{P}\in\mathscr{M}$, $Y|X=\mathbf{x}$ has a finite second moment for $\mathbb{P}$-almost every $\mathbf{x}$. \cite{williamson2021nonparametric} define a non parametric measure of variable importance.
\begin{equation}
    \psi_s(P) := \frac{\int\{\mu_P(\mathbf{x})-\mu_{P,s}(\mathbf{x})\}^2dP(\mathbf{x})}{var_P(\mathbf{y})}= \bigg[1-\frac{E_P\{\mathbf{y}-\mu_P(\mathbf{x})\}^{2}}{var_P(\mathbf{y})}\bigg]-\bigg[1-\frac{E_P\{\mathbf{y}-\mu_{P,s}(\mathbf{x})\}^{2}}{var_P(\mathbf{y})}\bigg]
\end{equation}

$\psi_{s}(P)$ measures the importance of variable $\mathbf{x}_{i}$ $\forall i \in s$
related to the output variable  $\mathbf{y}_{}$.

$\psi_{s}(P)$ is a nonparametric extension of the usual ANOVA-based variable importance measure.

By definition, $\psi_{0,s}$ =$\psi_{s}(P_{0})$ and $\psi_{0,s}\in [0,1]$. 

\textbf{Estimation of $\psi_{0,s}$}:

Let $\theta_{s,P}(\mathbf{x}) = \int\{\mu_P(\mathbf{x})-\mu_{P,s}(\mathbf{x})\}^2dP(\mathbf{x})$ then $\psi_s(P) = \frac{\theta_{s,P}(\mathbf{x})}{var_P(\mathbf{y})}$. Having the estimators $\hat{\mu} $ and $\hat{\mu}_s$ of $\mu_0$ and $\mu_{0,s}$, a natural estimator of $\psi_{0,s}$ is given by:   

\begin{equation}
    \hat{\psi}_{naive,s} = \frac{\hat{\theta}_{naive,s}}{var_{\mathbb{P}_n}(\mathbf{y})}
\end{equation}

Where $ \hat{\theta}_{naive,s} = \frac{1}{n}\sum_{i=1}^{n}\{\hat{\mu}(\mathbf{x}_i)-\hat{\mu}_s(\mathbf{x}_i)\}^2$ and $var_{\mathbb{P}_n}(\mathbf{y}) = \frac{1}{n}\sum_{i=1}^{n}(\mathbf{y}_i-\bar{\mathbf{y}}_n)^2$ 

The estimator $	\hat{\psi}_{naive,s}$ is in general highly biased in the way that its bias does not tend to zero quickly enough to ensure the consistency at rate $n^{-1/2}$. This problem is solved by considering the adjusted estimator of $\psi_{0,s}$ define by:

\begin{equation}
   \hat{\psi}_{n,s} = \hat{\psi}_{naive,s} + 2 \frac{\sum_{i=1}^{n}\{\mathbf{y}_i-\hat{\mu}(\mathbf{x}_i)\}\{\hat{\mu}(\mathbf{x}_i)-\hat{\mu}_s(\mathbf{x}_i)\}}{\sum_{i=1}^{n}(\mathbf{y}_i-\bar{\mathbf{y}}_n)^2} 
\end{equation}

Other VIMs are not accurate when it comes to dealing with correlated variables. To achieve this, \cite{owen2017shapley} proposed Shapley Values to quantify the population where the value function is the variance explained. But in most cases, the exact estimation of the Shapley value variable importance measure is computationally expensive. \cite{williamson2020efficient} illustrated a computationally efficient method to estimate and achieve valid statistical inference on the Shapley Population Variable Importance Measure (SPVIM). In their work, they proposed an estimator that converges at an asymptotically optimal rate based on randomly sampling.

Among the most used VIMs, we have the random forest VIMs developed by \cite{breiman2001random}. Random Forest  algorithms are extensively used for the purpose of interpretation of the results. Their popularity is due to their ability to handle high-dimensional data and take into account potential correlations that may exist between variables. The random forest also has a powerful product of variable importance measure. Throughout the random forest R package, \cite{liaw2002classification}  proposed two algorithms for calculating variable importance measures.
\cite{archer2008empirical} investigated the ability of RF variable importance measures to identify the true predictor among a large set of candidate predictors. They conducted an extensive simulation study using 20 levels of correlation among the predictor variables and 7 levels of association between the true predictor and the dichotomous response. As result, they conclude that When the study's goals are to produce an accurate classifier and to provide insight into the discriminative ability of individual predictor variables, the random forest methodology is appealing for use in classification problems. A common problem with the random forest variable important measure and also many other VIMs is that it cannot be used directly when there are missing values in the data. To overcome that, \cite{hapfelmeier2014new} has developed a new VIM method that can be applied directly to any category of data (data with or without missing value).

\cite{dong2019variable} presents the concept of a variable importance cloud, that maps each variable to its importance in every good predictive model. They demonstrate the variable importance cloud's properties and make connections to other areas of statistics. Variable importance diagrams are introduced as a two-dimensional projection of the variable importance cloud for display purposes.
\section{Mathematical background of the different methods}
\label{sec1:headings}
The five methods for estimating the importance of variables used in this paper for comparison purposes are: LASSO VIMs, RF VIMs, SVM VIMs, XGBOOST VIMs, and the PERF as VIMs. Through the following lines, we will illustrate how each of those techniques works mathematically.

\subsection{LASSO VIMs}
LASSO, standing for least absolute shrinkage and selection operator, is a regression analysis technique that can perform two main tasks which are: Variable selection and Regularization in order to improves the accuracy of prediction and the interpretability of the resulting model.
In this paper, we will focus only on the variable selection task.

Given a data matrix $\mathbf{X}$ and the response variable vector $\mathbf{y}$, the loss function of the LASSO regression is\begin{equation}\label{3.1.1}
    \hat{\beta}(\lambda) = \underset{\beta}{argmin} \big(\lVert \mathbf{y}-\mathbf{X}\beta\rVert^2+\lambda\lVert\beta\rVert_1\big)
\end{equation}
Where $\lVert\beta\rVert_1 = \sum_{j=1}^{p} \lvert\beta_j\rvert$ refer to $l^1$-norm and $\lambda>0$ is the parameter used to measure the strength of the penalty term. The additional term of \ref{3.1.1} is the regularized term used to avoid overfitting of the model.

The optimization problem \ref{3.1.1} is equivalent to the following optimization problem:

\begin{align}
    min \lVert \mathbf{y}-\mathbf{X}\beta\rVert^2 \text{  subject to  }  \lVert\beta\rVert_1<t
\end{align}

Where $t$ represents the upper bound of the sum of coefficients.

By taking the $l^2$-norm of the additional term of equation \ref{3.1.1} ie $\lVert\mathbf{\beta}\rVert_2^2 = \sum_{i=1}^{p}\beta_j^2$, we obtain the ridge regression.

\textbf{How LASSO access importance to variables?}

\begin{figure}[H]
    \centering
	\includegraphics[scale=0.5]{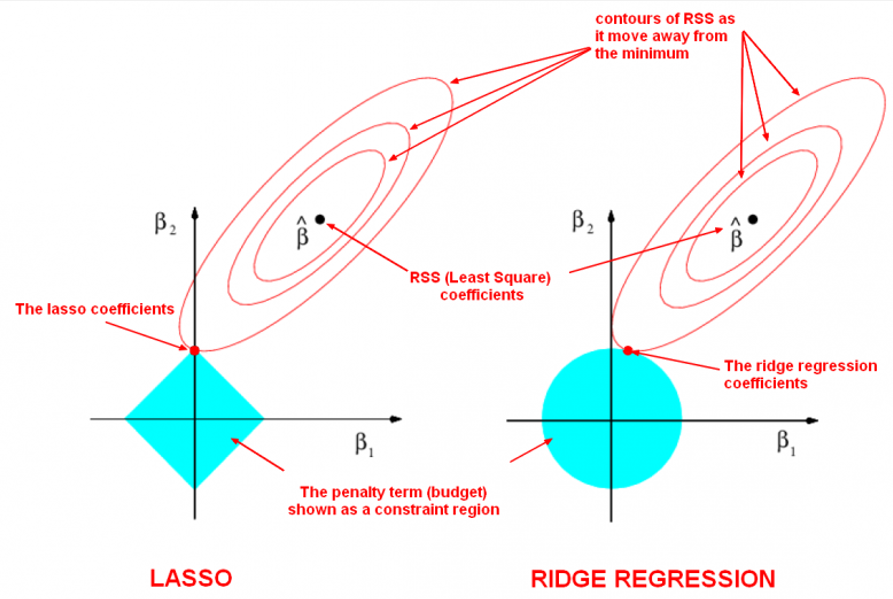}
	\caption{Image from \cite{bworld}}
	\label{louis}
\end{figure}

For the LASSO method, the constraint region is shaped like a diamond centered on zero. Since the objective function has an elliptic form, if a point results from the interception of the ellipse and a corner of the diamond, its coefficient $\beta_j$ is null then the variable $\mathbf{X}_j$ associated with that coefficient is not important for the LASSO model. In the case of ridge regression, the constraint region is a disk (In 2-dimension) means it doesn't have any corners. therefore, the coefficients are not necessarily equal to zero even if there is an interception between the disk and the ellipse. fig  \ref{louis}gives an illustration in 2D of this. That is the reason why Ridge regression cannot be used to select variables.  

\subsection{RF VIM}
Among the important product of the random forest, we have the random forest variable importance measure developed by \cite{breiman2001random}. In fact, we can use the random forest to evaluate the importance of variables either in a classification or regression task. The RF VIMs are done through two measures of significance which are the Mean Decrease Impurity (MDI) and the Mean Decrease Accuracy (MDA) see \cite{breiman2001random}.

The MDI can be calculated by averaging the overall decrease in node impurity caused by splitting the variable across all trees.

\textbf{Notations:}

We use the following notations:

\begin{itemize}
    \item $\mathbf{X}$ = $(X^{(1)},X^{(2)},\dots,X^{(p)})$ is our data input
    \item $M$ denote the number of overall trees
    \item $A$ is a generic cell and $N_n(A)$ is the number of data points falling in $A$
    \item $mtry\in\{1,2,\dots,p\}$ the number of possible split at each node of the tree
    \item $\mathcal{M}_{try}$ is the subset of selected coordinated after the split
    \item for $j = \{1,2,\dots,p\}$, we denote by the pair $(j,z)$ a cut in $A$ where $z$ is the position of the cut along the $j$-th coordinate. $\mathcal{C}_A$ is the set of all such possible cut in $A$.
\end{itemize}

For a variable $X^{(j)}$ it MDI is define by see: \cite{biau2016random}
\begin{align}\label{3.4.1}
    \hat{MDI}(X^{(j)}) = \frac{1}{M}\sum_{l=1}^{M}\sum_{\overset{t\in\mathcal{T}_l}{j_{n,t}^* = j}}p_{n,t}L_{reg,n}(j_{n,t}^*,z_{n,t}^*)
\end{align}

Where $p_{n,t}$ is the proportion of observations falling in the node $t$,  $\{\mathcal{T}_l\}_{1\leq l\leq M}$ are all the trees we have in the forest, $\displaystyle(j_{n,t}^*,z_{n,t}^*) \in \underset{(j,z)\in\mathcal{C}_A}{\underset{j\in\mathcal{M}_{try}}{\text{arg max }}}L_{reg,n}(j,z)$ and finally, $L_{reg,n}(j,z)$ is known as the regression CART (classification and regression trees) split criterion define by. 

\begin{align}
    L_{reg,n}(j,z) = \frac{1}{N_n(A)}\sum_{i = 1}^{n}(Y_i-\overline{Y}_A)^2\mathbbm{1}_{X_i\in A} - \frac{1}{N_n(A)}\sum_{i = 1}^{n}(Y_i-\overline{Y}_{A_{L}}\mathbbm{1}_{X_i^{(j)}<z}-\overline{Y}_{A_{R}}\mathbbm{1}_{X_i^{(j)}\geq z})^2\mathbbm{1}_{X_i\in A}
\end{align}
 
with $A_L = \{x\in A \text{ such that } X^{(j)}<z \}$,  $A_R = \{x\in A \text{ such that } X^{(j)} \geq z\}$ and $\overline{Y}_{A_{L}}$, $\overline{Y}_{A_{R}}$, $\overline{Y}_{A_{}}$ are respectively the average of all the $Y_i$ belonging to  $A_L$, $A_R$, $A$.

The formula \ref{3.4.1} is valid only for the random forest regression. For the classification, instead of $L_{reg,n}(j,z)$, we use $L_{class,n}(j,z)$ define by:
\begin{align*}
    L_{class,n}(j,z) = p_{0,n}(A) p_{1,n}(A) - \frac{N_n(A_L)}{N_n(A)}\times p_{0,n}(A_L)p_{1,n}(A_L) - \frac{N_n(A_R)}{N_n(A)}\times p_{0,n}(A_R)p_{1,n}(A_R)
\end{align*}
Where $p_{0,n}(A)$ and $p_{1,n}(A)$ are the empirical probability of a data points in the cell $A$ having label 0 and 1.

As a result, the MDI of $X^{(j)}$  calculates the weighted decrease in impurity relating to splits along $X^{(j)}$ and average it across all trees.

When coming to the MDA, is based on a different approach and uses the OOB error estimation to measure the importance of a variable. We randomly permute the values of the $j$-th variable $X^{(j)}$ in the OOB observation and place these examples down in the tree to determine their importance. 

The MDA of the variable $X^{(j)}$ is calculated by taking the average of the difference in OOB error estimation before and after the permutation across all the trees and it's defined by: see \cite{biau2016random}

\begin{align}
    \hat{MDA}(X^{(j)}) = \frac{1}{M}\sum_{l = 1`}^{M}\Bigg[R_n\big[m_n(,\Theta_l),D_{l,n}^j] - R_n\big[m_n(,\Theta_l),D_{l,n}\big]\bigg]
\end{align}

Where for $D = D_{l,n}$ or $D = D_{l,n}^j$, we have $\displaystyle R_n\big[m_n(,\Theta_l),D_{}] = \frac{1}{|D|}\sum_{i,(X_i,Y_i)\in D}^{}(Y_i - m_n(X_i,\Theta_l))^2 $ and $m_n(X_i,\Theta_l)$ is the $l$-th tree estimate ie $m_n(X_i,\Theta_l) = \hat{\mathbb{E}(Y| X = X_i)}$

\subsection{Support vector machines vaiable importance}
\subsubsection{An introduction to support vector machine}
Developed by \cite{vapnik1963pattern}, support vector machines (SVM) are among the most powerful predictions method. They are  supervised learning models and can be performed to analyze data for either classification or regression analysis. Support vector machines can be divided into two groups:
\begin{itemize}
    \item[-] Linear SVM
    \item[-] Nonlinear SVM
\end{itemize}
We are in the presence of linear SVM when we can use a hyperplane (kernel of linear form e.g.: straight line) to separate classes in the data. When this is not possible ie we cannot linearly separate the classes into the data, we have a nonlinear SVM. Mathematically, linear SVM can be modelled by $y = <\mathbf{w},\mathbf{x}>+b$ and for non linear SVM, we have $y = <\mathbf{w},\phi(\mathbf{x})>+b$ with $\mathbf{w},\mathbf{x}\in\mathbb{R}^n$, $b\in\mathbb{R}$.

For the case of classification, we consider the two classes $\mathcal{D}^+$ and $\mathcal{D}^-$ such that:

$\mathcal{D}^+ = \{\mathbf{x}_i\in\mathbb{R}^n, \mathbf{w}^T\mathbf{x}_i-b\geq 1\}$

$\mathcal{D}^- = \{\mathbf{x}_i\in\mathbb{R}^n, \mathbf{w}^T\mathbf{x}_i-b\leq -1\}$

$y_i = 1$ if  $\mathbf{x}_i\in\mathcal{D}^+$  and   $y_i = -1$  if $\mathbf{x}_i\in\mathcal{D}^-$

ie \begin{align*}
    \begin{cases}
   y_i = 1   \text{ if } \mathbf{w}^T\mathbf{x}_i - b \geq 1\\
  y_i = -1    \text{ if } \mathbf{w}^T\mathbf{x}_i - b \leq -1
     \end{cases}\implies y_i(\mathbf{w}^T\mathbf{x}_i - b)\geq 1 \quad\forall \mathbf{x}_i\in\mathbb{R}^n
\end{align*}

\begin{figure}[H]
    \centering
	\includegraphics[scale=0.5]{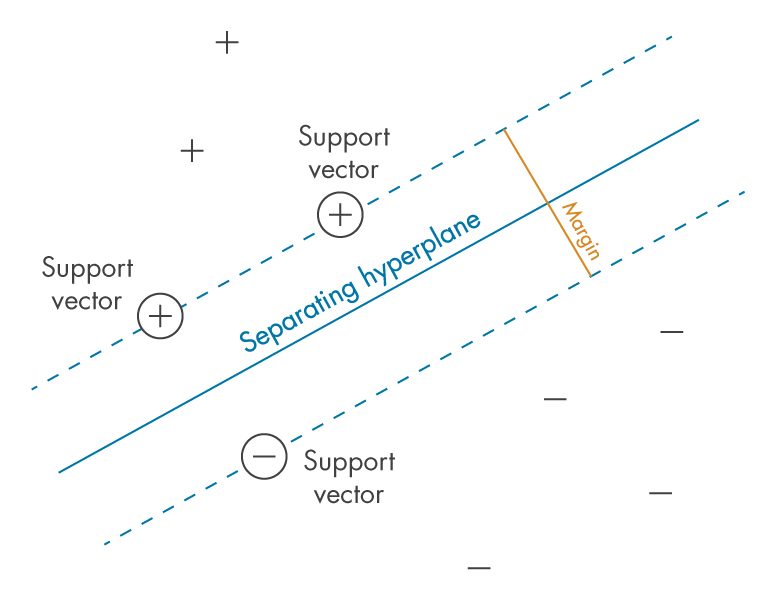}
	\caption{An illustration of SVM}
	\label{3.3}
\end{figure}

The idea of SVM is to find the margin that maximizes the two classes $\mathcal{D}^+$ and $\mathcal{D}^-$ and then, the optimal hyperplane is the hyperplane that passes in the middle of the maximum margin.   
We have an illustration in fig \ref{3.3}

Geometrically, $d(\mathcal{D}^+,\mathcal{D}^-) = d(A,B)  =  \frac{2}{\sqrt{\lVert \mathbf{w} \rVert^2+1}}$ where:

$A = \{\mathbf{x}_i\in\mathbb{R}^n, \mathbf{w}^T\mathbf{x}_i-b = 1\}$

$B = \{\mathbf{x}_i\in\mathbb{R}^n, \mathbf{w}^T\mathbf{x}_i-b = -1\}$

To maximize that distance, it is the same to maximize : $\underset{s.t. y_i(\mathbf{w}^T\mathbf{x}_i-b)\geq1}{\underset{\mathbf{w},b}{Max}\quad\frac{2}{\lVert \mathbf{w} \rVert^{2}}}$ which is equivalent to the following optimization problem: 

\begin{align}\label{3.5.1}
    \underset{s.t. y_i(\mathbf{w}^T\mathbf{x}_i-b)\geq1}{\underset{\mathbf{w},b}{Min} \quad\frac{1}{2}\lVert \mathbf{w} \rVert^{2}}\quad\forall i \in \{1,2,\dots,n\}
\end{align}

To solve the optimization problem in \ref{3.5.1}, we use the lagrangian $L(\mathbf{w},b,\lambda)$ define as:

\begin{align*}
    L(\mathbf{w},b,\lambda) = \frac{1}{2}\lVert \mathbf{w} \rVert^{2} - \sum_{i=1}^{n}\lambda_i(y_i(\mathbf{w}^T\mathbf{x}_i-b)-1)
\end{align*}
with $\lambda = (\lambda_1,\dots,\lambda_n)$.

Applying the  Karush–Kuhn–Tucker (KKT) conditions yield:

\begin{align*}
	\begin{cases}
		\frac{\partial L(\mathbf{w},b,\lambda)}{\partial \mathbf{w}} =\mathbf{w} - \displaystyle\sum_{i=1}^{n}\lambda_iy_i\mathbf{x}_i = 0\\\\
		\frac{\partial L(\mathbf{w},b,\lambda)}{\partial b} = \displaystyle\sum_{i=1}^{n}\lambda_iy_i = 0\\\\
		\frac{\partial L(\mathbf{w},b,\lambda)}{\partial \lambda_i} = y_i(\mathbf{w}^T\mathbf{x}_i - b) = 1
	\end{cases} \implies \begin{cases}
	\mathbf{w} = \displaystyle \sum_{i=1}^{n}\lambda_iy_i\mathbf{x}_i \\\\
	\displaystyle\sum_{i=1}^{n}\lambda_iy_i = 0\\\\
	 y_i(\mathbf{w}^T\mathbf{x}_i - b) = 1
\end{cases}
\end{align*}

Substituting the value of $\mathbf{w}$ into the lagrangian yield to:

\begin{align*}
	L(w,b,\lambda)& = \frac{1}{2}\bigg(\displaystyle \sum_{i=1}^{n}\lambda_iy_i\mathbf{x}_i\bigg)^T\bigg(\displaystyle \sum_{i=1}^{n}\lambda_iy_i\mathbf{x}_i\bigg)-\sum_{i=1}^{n}\lambda_i\bigg[y_i\bigg(\big(\displaystyle \sum_{j=1}^{n}\lambda_jy_j\mathbf{x}_j\big)^T\mathbf{x}_i-b\bigg)-1\bigg]\\
	& = \frac{1}{2}\sum_{i,j=1}^{n}\lambda_i\lambda_jy_iy_j\mathbf{x}_i^T\mathbf{x}_j-\sum_{i,j=1}^{n}\lambda_i\lambda_jy_iy_j\mathbf{x}_i^T\mathbf{x}_j+\underbrace{\sum_{i=1}^{n}\lambda_iy_ib}_{=0}+\sum_{i=1}^{n}\lambda_i\\
	& = -\frac{1}{2}\sum_{i,j=1}^{n}\lambda_i\lambda_jy_iy_j\mathbf{x}_i^T\mathbf{x}_j+\sum_{i=1}^{n}\lambda_i
\end{align*}

The dual form of the optimization problem \ref{3.5.1} is given by:
\begin{align}\label{3.5.2}
	&\underset{\lambda}{\text{Max }} L(\mathbf{\mathbf{w}},b,\lambda) = -\frac{1}{2}\sum_{i,j=1}^{n}\lambda_i\lambda_jy_iy_j\mathbf{x}_i^T\mathbf{x}_j+\sum_{i=1}^{n}\lambda_i\\
	& \text{with } \lambda_i\geq0\quad\forall i\in\{1,\dots,n\}, \sum_{i=1}^{n}\lambda_iy_i=0\nonumber
\end{align}

ie \begin{align}\label{3.5.3}
	 L(\mathbf{w},b,\lambda)& = \sum_{i=1}^{n}\lambda_i-\frac{1}{2}\sum_{i,j=1}^{n}\lambda_i\lambda_jy_iy_j\mathbf{x}_i^T\mathbf{x}_j\\
	& = \sum_{i=1}^{n}\lambda_i-\frac{\lambda^T\lambda}{2}\underbrace{\sum_{i,j=1}^{n}k(i,j)}_{K}\text{   With $ k(i,j) = y_jy_i\mathbf{x}_j^T\mathbf{x}_i $}\nonumber\\
	& = \sum_{i=1}^{n}\lambda_i - \frac{\lambda^T\lambda}{2}K\nonumber
\end{align}
The optimization problem in \ref{3.5.2} can be then rewrite as:

\begin{align}\label{3.5.4}
	&\underset{\lambda}{\text{Max }} L(\mathbf{w},b,\lambda) =  \sum_{i=1}^{n}\lambda_i - \frac{\lambda^T\lambda}{2}K\\
	& \text{with } \lambda_i\geq0\quad\forall i\in\{1,\dots,n\}, \sum_{i=1}^{n}\lambda_iy_i=0\nonumber
\end{align}

Equation \ref{3.5.4} is the final optimization problem where we obtain the value of $\lambda$. After having the value of $\lambda$, we can get the optimum value of $\mathbf{w}$ from $\mathbf{w} = \displaystyle\sum_{i=1}^{n}\lambda_iy_i\mathbf{x}_i$.

Let us now find the optimum $b$. For a given support vector $\mathbf{x}_i$, we have (from the KKT result):
\begin{align*}
	y_i(\mathbf{w}^T\mathbf{x}_i-b) = 1&\implies y_i(\sum_{j=1}^{n}\lambda_jy_j\mathbf{x}_j^T\mathbf{x}_i-b) = 1\\
	&\implies y_i^2(\sum_{j=1}^{n}\lambda_jy_j\mathbf{x}_j^T\mathbf{x}_i-b) = y_i\quad\text{We know that $y_i^2 = (\pm1)^2 = 1$}\\
	&\implies b = y_i - \sum_{j=1}^{n}\lambda_jy_j\mathbf{x}_j^T\mathbf{x}_i
\end{align*}
Instead of taking a random support vector, we can take the average over the set of all support vector $S$. We then get:
\begin{align*}
	b = \frac{\displaystyle\sum_{i \in S}(y_i-\sum_{j=1}^{n}\lambda_jy_j\mathbf{x}_j^T\mathbf{x}_i)}{\lvert S\rvert}
\end{align*}

with those value of $\mathbf{w}$ and $b $ our optimum classifier is then $f(\mathbf{x}) = sgn(\mathbf{w}^T\mathbf{x}-b)$ where $sgn$ is define by: 

\begin{align*}
    sgn(x) = \begin{cases}
    1\text{ if } x>0\\
    0\text{ if } x=0\\
    -1\text{ if } x<0
    \end{cases}
\end{align*}

In the case of non linear SVM classifier,  an extra set of variables $\epsilon_i$ is introduced to weight the cost of miss classification. The optimization problem is now define as:
\begin{align*}
   & \underset{\mathbf{w},b,\epsilon}{Min}\quad \frac{1}{2}\lVert \mathbf{w}\rvert^2+ c\sum_{i=1}^{m}\epsilon_i\\
   & \text{s.t. } y_i(\mathbf{w}^T\phi(\mathbf{x}_i)+b)\geq 1-\epsilon_i\quad \forall i = 1,2,\dots,n\\
   & \quad\quad \quad\quad\epsilon_i\geq0 \quad \forall i = 1,2,\dots,n
\end{align*}

Where $\phi:\mathbf{x}\mapsto \phi(\mathbf{x})$ is a nonlinear function that maps the training data to the higher dimensional space $\mathscr{Y}$ and $c$ is a penalty parameter.

Solving by using the same techniques as above give the solution of the optimization problem: $f(\mathbf{x}) = sgn(\mathbf{w}^T\phi(\mathbf{x})+b)$.

 \subsubsection{Support vector machines VIMs}
 
For the SVM, it exists three mains direction of estimating the importance of variables. We have The filter, the wrapper, and the embedded methods But we focussed only on the wrapper method.

A wrapper approach searches the feature space for feature subsets with the highest predictive power, then optimizes the induction method that uses the subset for classification (see \cite{maldonado2009wrapper}). Although the wrapper method is computationally extensive, it produces more accurate results than filter techniques in many cases.

Having a set of $n$ variables, the wrapper method works by attempting to select $m$ features ($m<n$) that result in the greatest class separation margin. Since those $m$ variables are the one used in the final decision rule, the method attempt  to find the best subset of the $m$ variables.

This problem was solved by sequentially deleting one feature at a time until only $m$ are left we call this process backward variable selection.

We define \begin{align}
    W^2(\lambda) = \sum_{i=1}^{n}\sum_{j=1}^{n}\lambda_i\lambda_jy_iy_jK(\mathbf{x}_i,\mathbf{x}_j)
\end{align}

For a feature $p$, we define 
\begin{align}
    W_{(-p)}^2(\lambda) = \sum_{i=1}^{n}\sum_{j=1}^{n}\lambda_i\lambda_jy_iy_jK(\mathbf{x}_i^{(-p)},\mathbf{x}_j^{(-p)})
\end{align}
Where $\mathbf{x}_i^{(-p)}$ means training feature $i$ without the feature $p$ and $K$ is a given kernel. To select variables, the wrapper method consists to remove the feature with the smallest value of  $\lvert W^2(\lambda)-W_{(-p)}^2(\lambda)\rvert$

\subsubsection{An introduction to XGBOOST}

XGBOOST (Extreme Gradient Boosting) is among the most popular machine learning techniques used in many applications such as Fraud detection, image classification, etc... XGBOOST can handle classification, regression, and ranking tasks. Due to its wide variety of applications, XGBOOST during these past few years has gained significant attention and has helped many people to win some machine learning competitions like the ones regularly organized by Kaggle. 

What is impressive with XGBOOST is its scalability. In fact, on a single machine, XGBOOST can run more than ten times faster than the existing popular machine learning algorithm \cite{priscilla2020influence}. XGBOOST's scalability can be explained by the number of important systems and algorithmic processes of optimization.

\subsection{Mathematical modeling of XGBOOST}

Let consider the data structure below:

\begin{align*}
    \mathscr{D}_{n} = \{(\mathbf{x}_i,y_i), \mathbf{x}_i\in\mathbb{R}^m,y_i\in\mathbb{R}, i=1,2,\dots,n\}
\end{align*}

We design an ensemble model of $K$ trees. The response variable $y_i$ is then predicted as:

\begin{align*}
    \hat{y}_i = \phi(\mathbf{x}_i) = \sum_{k=1}^{K}f_k(\mathbf{x}_i),\qquad f_k\in\mathcal{F}
\end{align*}

where $f_k$ for $k=1,\dots,n$ is the functional space of each tree. $\mathcal{F}$ is the collection of all the possible classification and regression trees (CART).

The objectives function to be minimized is defined by:

\begin{align}\label{a}
    \mathcal{L} = \sum_{i=1}^{n}l(y_i,\hat{y}_i)+ \sum_{k=1}^{K}\Omega(f_k)
\end{align}

where $l$ is a convex loss function  and the second term of the addition is the regularization term used to avoid overfitting by smoothing the final learn weight \cite{chen2016xgboost}.

Since the objectives function in equation \ref{a} have a function as a parameter, it can't be optimized by using traditional methods of optimization. Therefore, the model has to be trained differently.

Let $\hat{y}_i^{(t)}$ the prediction at the $t$-th iteration of the $i$-th variable. We have the following recursive formula:
\begin{align*}
    \hat{y}_i^{(t)} = \hat{y}_i^{(t-1)}+f_t(\mathbf{x}_i)
\end{align*}
Then the objective fumction is now:
\begin{align*}
    \mathcal{L}^{(t)} = \sum_{i=1}^{n}l(y_i,\hat{y}_i^{(t-1)}+f_t(\mathbf{x}_i))+\Omega(f_t)
\end{align*}

We can use the 2nd order Taylor expansion to give an approximation of $\mathcal{L}^{(t)}$ ie
\begin{align}\label{b}
    \mathcal{L}^{(t)}\approx \sum_{i=1}^{n}\Bigg[l\big(y_i,\hat{y}_i^{(t-1)}+g_if_t(\mathbf{x}_i))+\frac{1}{2}h_if_t^2(\mathbf{x}_i)\big)\bigg]+\Omega(f_t)
\end{align}
where $g_i = \frac{\partial l(y_i,\hat{y}_i^{(t-1)})}{\partial \hat{y}^{(t-1)}}$ and $h_i = \frac{\partial^2 l(y_i,\hat{y}_i^{(t-1)})}{\partial \hat{y}^{(t-1)^2}}$

\textbf{NB:} Our goal is to find $f_t$ that minimize \ref{b}

In equation \ref{b}, $l\big(y_i,\hat{y}_i^{(t-1)}\big)$ is a constant term since it does not depend on $f_t(\mathbf{x}_i)$ then it can be removed we then get the new objective function at step $t$ as:
\begin{align}\label{c}
    \mathcal{L}_{new}^{(t)}  = \sum_{i=1}^{n}\big[g_if_t(\mathbf{x}_i)+\frac{1}{2}h_if_t^2(\mathbf{x}_i)\big]+\Omega(f_t)
\end{align}
let define $I_j = \{i \text{ s.t. } q(\mathbf{x}_i) = j \}$ where $q$ is the function that has been assigned to each data point of the corresponding leaf ie :
        \begin{align*}
            &q : \mathbb{R}^d\to\{1,2,\dots,T\}\\
            & \mathbf{w}\in\mathbb{R}^T \text{ is the leaf weight of the tree }\\
            & f_t(\mathbf{x}) = w_{q(\mathbf{x})}
        \end{align*}
Also, we define the complexity of the tree $t$ as: \begin{align*}
    \Omega(f) = \gamma T + \frac{1}{2}\lambda\sum_{j=1}^{T}w_j^2
\end{align*} where $w_j$ is the score of the $j$-th leaf; $T$ is the total number of leaves or terminal nodes and $\gamma$ is the penalty definable by the user. (This defines the complexity of the tree $t$ and we can have many possible definitions)

Replacing this into equation \ref{c} yield to:

\begin{align}\label{d}
	\mathcal{L}_{new}^{(t)}& = \sum_{i=1}^{n}\big[g_if_t(\mathbf{x}_i)+\frac{1}{2}h_if_t^2(\mathbf{x}_i)\big] + \gamma T + \frac{1}{2}\lambda\sum_{j=1}^{T}w_j^2\nonumber\\
	& = \sum_{i=1}^{n}\big[g_iw_{q(\mathbf{x}_i)}+\frac{1}{2}h_iw_{q(\mathbf{x}_i)^2}\big]+\gamma T+\frac{1}{2}\lambda\sum_{j=1}^{T}w_j^2\nonumber\\
	& = \sum_{j=1}^{T}\Bigg[\bigg(\sum_{i\in I_j}g_i\bigg)w_j+\frac{1}{2}\bigg(\sum_{i\in I_j}h_i+\lambda\bigg)w_j^2\Bigg]+\gamma T
\end{align}

For a leaf $j$, we want $\frac{d \mathcal{L}_{new}^{(t)}}{d w_j} = 0$ ie  $\sum_{i\in I_j}g_i + (\sum_{i\in I_j}h_i+\lambda)w_j = 0$. Then, the optimal value of $w_j$ of the leaf $j$ is 

\begin{align*}
    w_j^* = -\frac{\sum_{i\in I_j}g_i}{\sum_{i\in I_j}h_i+\lambda}
\end{align*}

Substituting this back into equation \ref{d} give:

\begin{align}
    \mathcal{L}_{new}^{(t)} = -\frac{1}{2}\sum_{j=1}^{T}\frac{\bigg(\sum_{i\in I_j}g_i\bigg)^2}{\sum_{i\in I_j}h_i+\lambda}+\gamma T
\end{align}

For a given base learner of $T$ nodes, $ \mathcal{L}_{new}^{(t)}$ found above is the best loss.

Let $I_L$ and $I_R$ the left and right node after splitting the node $t$ and $I = I_L \bigcup I_R$ then the lost reduction after split is:

\begin{align}
    \mathcal{L}_{split}^{(t)} =   \mathcal{L}_{new,right}^{(t)}+\mathcal{L}_{new,left}^{(t)}-\mathcal{L}_{new}^{(t)}
\end{align}

Since we have only one node, $T = 1$. Then 
    \begin{align*}
	\mathcal{L}_{split}^{(t)}& = \frac{1}{2}\frac{\big(\sum_{i\in I_L}g_i\big)^2}{\sum_{i\in I_l}h_i+\lambda}+\gamma\times 1 +  \frac{1}{2}\frac{\big(\sum_{i\in I_R}g_i\big)^2}{\sum_{i\in I_R}h_i+\lambda}+\gamma\times 1 - \frac{1}{2}\frac{\big(\sum_{i\in I}g_i\big)^2}{\sum_{i\in I}h_i+\lambda}-\gamma\times 1\\
	& = \frac{1}{2}\bigg[\frac{\big(\sum_{i\in I_L}g_i\big)^2}{\sum_{i\in I_l}h_i+\lambda}+\frac{\big(\sum_{i\in I_R}g_i\big)^2}{\sum_{i\in I_R}h_i+\lambda}- \frac{\big(\sum_{i\in I}g_i\big)^2}{\sum_{i\in I}h_i+\lambda}\bigg]+\gamma
\end{align*}

\subsubsection{XGBOOST VIMs}
An advantage of XGBOOST is that once an XGBOOST model is trained, it automatically computes the score of the features used in the model.

\textbf{How does XGBOOST measure the importance of variables?:}

In the case of XGBOOST, the score attributed to each variable was estimated in almost the same way as in the case of the random forest variable importance \cite{sandri2008bias}. Therefore we will use the same notation as in the case of random forest variable importance. In fact, the importance of a variable $X_j$ was calculated as:  

\begin{align}\label{e}
    \hat{VI}(X^{(j)}) = \frac{1}{M}\sum_{l=1}^{M}\sum_{\overset{t\in\mathcal{T}_l}{j_{n,t}^* = j}}p_{n,t}L_{reg,n}^{2}(j_{n,t}^*,z_{n,t}^*)
\end{align}

the formula in \ref{e} is valid only in the case of regression, for classification, we just replace $L_{reg,n}$ by $L_{class,n}$.

\subsection{PERF as VIMs}

In the context of ensemble learning, \cite{fokoue2015prediction} developed a novel variable importance score named $PERF(\cdot)$ standing for the predictive error function as VIM. As RF, SVM and XGBOOST, $PERF$ can handle both regression and classification tasks. Contrary to XGBOOST and RF, $PERF$ can be performed beyond ensemble trees means the base learner can be made by any machine learning techniques like multiple linear regression, SVM, and logistic regression either inhomogeneous or heterogeneous ensemble. 

\textbf{How does PERF work?}

We denote the dataset by: \begin{align*}
    \mathscr{D}_n = \{(\mathbf{x}_i,y_i), \mathbf{x}_i\in\mathbb{R}^p, y_i\in\mathscr{Y}, i\in\{1,2,\dots,n\}\}
\end{align*}

$\mathscr{Y} = \mathbb{R}$ in case of regression and $\mathscr{Y} = \{1,2,\dots,K\}$ for the case of classification.

We have \begin{align*}
   f:\text{  }  &\mathbb{R}^p \to \mathscr{Y}\\
    &     \mathbf{x}_i \mapsto y_i
\end{align*}
The unknown function that match properly the data. $\hat{f}$ is the estimator of $f$ that minimize the loss function $l(\cdot,\cdot)$ define as:
\begin{align*}
    l(y_j,\hat{f}(\mathbf{x}_j))& = \mathbbm{1}_{y_j\neq\hat{f}(\mathbf{x}_j)}\\
    & = \begin{cases}
        1 \text{ if } y_j\neq\hat{f}(\mathbf{x}_j)\\
        0 \text{ else }
    \end{cases}
\end{align*}

for classification task and we can take the $l^2$ loss for regression ie 
\begin{align*}
   l(y_j,\hat{f}(\mathbf{x}_j)) = \lVert y_j - \hat{f}(\mathbf{x}_j) \rVert^2
\end{align*}

Let consider $\gamma = (\gamma_1,\dots,\gamma_p)^T$ a $p$ dimensional vector of indicator function s.t.
\begin{align*}
    \gamma_j = \mathbbm{1}_{ \text{\{ $\mathbf{x}_j$ is active in the model indexed by $\gamma$}\}}
\end{align*}

Having an ensemble of $B$ models\begin{align*}
    \mathscr{H} = \{h(\cdot,\gamma^{(i)})\}_{i = 1,\dots,B}
\end{align*}

Where $h(\cdot,\gamma^{(i)})$ is the built function with active variables in the $i$-th model of the ensemble and $\gamma^{(i)} = (\gamma_1^{(i)},\dots,\gamma_p^{(i)})$ with:\begin{align*}
    \gamma_j^{(i)} = \mathbbm{1}_{ \text{\{ $\mathbf{x}_j$ is active in the $i$-th model of the ensemble}\}}
\end{align*}

The importance of the variable $\mathbf{x}_j        $ was calculated using $PERF$ as \cite{fokoue2015prediction}:
\begin{align*}
	PERF(\mathbf{x}_j)& = \frac{1}{B} \sum_{b=1}^{B}score(h(\cdot,\gamma^{(b)}))-\frac{1}{B_j}\sum_{b=1}^{B}\gamma_j^{(b)}score(h(\cdot,\gamma^{(b)}))\\
	& = \text{Average score over all models - Average score over all models with $\mathbf{x}_j$}
\end{align*}
Where $B_j$ denotes the number of models using feature $\mathbf{x}_j$: ie

\begin{align*}
		B_j = \sum_{b=1}^{B}\mathbbm{1}_\{{\gamma_j^{(b)}=1\}}
\end{align*}

		$B$ is the total number of model in the ensemble and \begin{align*}
			score(h(\cdot,\gamma^{(b)})) = \frac{1}{\lvert \mathscr{D}_n^{(b)}\rvert}\sum_{\mathbf{x}_i\notin\mathscr{D}_n^{(b)}}l(y_i,h(\mathbf{x}_i,\gamma^{(b)}))
		\end{align*}
		
		$\mathscr{D}_n^{(b)}$ is the sample drawed from $\mathscr{D}_n$ used by the learner $h(\cdot,\gamma^{(b)})$ therefore, a naive estimation of PERF is:
		\begin{align*}
			\hat{PERF}(\mathbf{x}_j)& = \frac{1}{B} \sum_{b=1}^{B}score(\hat{h}(\cdot,\gamma^{(b)}))-\frac{1}{B_j}\sum_{b=1}^{B}\gamma_j^{(b)}score(\hat{h}(\cdot,\gamma^{(b)}))\\
			& = \text{Average score over all models - Average score over all models with $\mathbf{x}_j$}
		\end{align*}

\textbf{Some properties of PERF:}

For two given variables $\mathbf{X}_j$ and $\mathbf{X}_k$, 
\begin{itemize}
    \item If $PERF(\mathbf{X}_j)> PERF(\mathbf{X}_k)$, then variable $\mathbf{X}_j$ is more important than variable $\mathbf{X}_k$.
    \item If  $PERF(\mathbf{X}_j) =  PERF(\mathbf{X}_k)$, then the two variables have the same importance.
    \item If $PERF(\mathbf{X}_j)\leq0$, then variable $\mathbf{X}_j$ is not useful and can then be withdraw from the model. 
\end{itemize}

\section{Experimentation design and implementation of the different method}
\label{sec2:headings}

To compare and measure the usefulness of each of the variable importance measures described in section 3, we use both simulated and real-life data in the context of classification and regression. Many scenarios have been considered: High correlated data set, ultrahigh dimensional data, and high redundancy data set.

\subsection{Variable importance score for simulated data with high correlation among the features}

\begin{figure}[H]
    \centering
	\includegraphics[height=10cm,width=10cm]{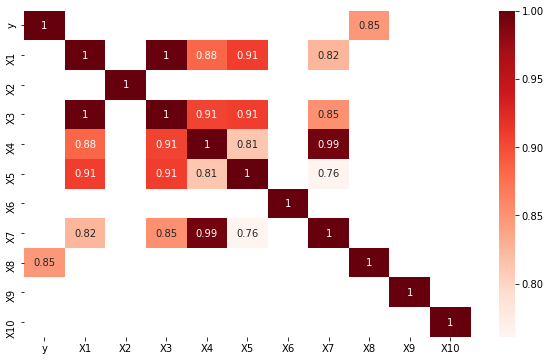}
	\caption{Heatmap of the simulated data with correlated features }
	\label{fig 4.1}
\end{figure}

Deal with data containing high correlated features is the nightmare of most variable importance estimation in our case, we simulated a data of 100 samples and 10 features. The true underlined function we defined is \begin{align*}
     f(X) = 1+20X_1+30X_2+9X_3+100X_8+40X_7\quad\text{such that } X_i\sim\mathcal{N}(0,1)\forall i\in\{1,2,5,6,8,9,10\}
\end{align*} 
 We adjusted the other features such that they can be highly correlated together as: \begin{align*}
    \begin{cases}
    X_3 = 0.1X_1+0.025X_2\\
    X_4 = 0.4X_3+0.1X_2\\
    X_7 = 0.6X_3+0.2X_2
    \end{cases}
\end{align*}this generates the heatmap in Figure \ref{fig 4.1} (\textbf{NB:} in the heatmap, we just showed features with a correlation higher than 0.7):

After computing the variable's importance scores of the five methods we are interested in, we plotted them in figure \ref{fig 4.2} and also we selected the most important features of each of the VIMs to train them on a neutral model (multiple linear regression) since we are in a regression case by taking each of the most important features of the VIMs method and all the features to train different multiple linear regression. We computed then three different errors: Mean Square Error(MSE), Mean Absolute Error, and the Root Mean Square Error(RMSE). The errors from the different models are stored in Table  \ref{table 4.2}. 

\subsection{Variable importance score for simulated data with redundant features}
In this section, we simulated data having a sample size of 1000 and ten features. Among the ten features, we made five redundant features by using the very nice function $make_{-}classification$ in the sklearn library on python. After computing and plotting the importance of variables in this simulated data, (see figure \ref{fig 4.3}) we used the most important features of each technique and trained a neutral model with those features. Since it's the case of classification, we chose logistic regression as our neutral model. We calculated the three different indicators of the performance of the logistic regression: The accuracy, the recall, and the precision for each of the logistic models built using the most important variables of the five techniques used and we store the result in table \ref{table}.

\subsection{Variable importance score on a real-life dataset: the gifted dataset}

The gifted dataset is a kind of infra high dataset ( $n = 38$ and $p = 8$). It has been made when an investigator decides to understand if there is any relationship between the analytical skills of young gifted children and the following variables: 
\begin{itemize}
    \item [-] fatheriq : the father's IQ
    \item[-] motheriq: Mother's IQ
    \item[-] speak: Age (in months) when the child counted to 10 successfully
    \item[-] read: Average number of hours the mother or father reads to the child per week
    \item[-] edutv: Average number of hours per week the child watched an educational program during the past three months.
    \item[-] cartoons: Average number of hours per week the child watched cartoons during the past three months.
    \item[-] score: score in test of analytical skills used as the response variable.
\end{itemize}
The score of each of those variables was plotted in figure \ref{fig 4.4}. To decide which VIMs got the right most important features, we preceded them in the same way as in section 4.1. the result errors from the different models were shown in the table \ref{table 4.1}.

\subsection{Variable importance score for an ultra-high dimension data: Prostate-Cancer dataset}

The prostate-cancer dataset is a very nice example of data where we can apply variable selection and then variable importance estimation. The data is mostly used in the field of biology and it has been collected on different patients to identify if they have prostate or not. Since the data contains 79 rows against 500 columns, it is an ultrahigh dataset. The response variable $y$ has two possible values 0 or 1. We have 1 when the patient is positive for the prostate and 0 otherwise.  We performed variables importance on this data using the five VIMs and the result is shown in Figure \ref{fig 4.5}. Since we have a lot of features, instead of putting the name of the variables on the y label, we denoted them by their index in the data; only XGBOOST showed the name of the features and their respective score. According to the complexity and the high number of features of the data, we also compute the running time of each method to see their computational time and we stored the results of each method in Table \ref{table 4.4}.

\subsection{Visualisation of the results}

\newpage
\begin{figure}[H]

	\centering
	
	\begin{subfigure}[b]{0.4\textwidth}
	\centering
    \includegraphics[scale=0.6]{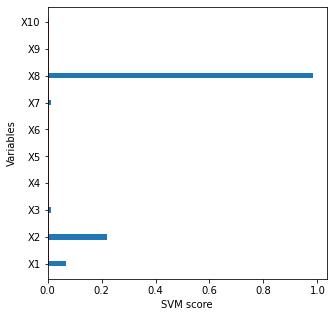}
	\end{subfigure}
	\begin{subfigure}[b]{0.4\textwidth}
		\centering
		\includegraphics[scale=0.6]{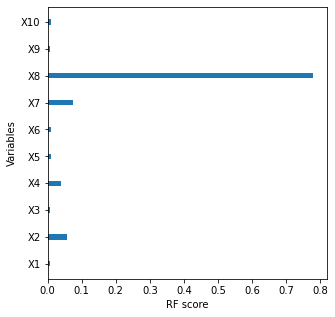}
	\end{subfigure}

	\centering
	\begin{subfigure}[b]{0.4\textwidth}
		\centering
		\includegraphics[scale=0.6]{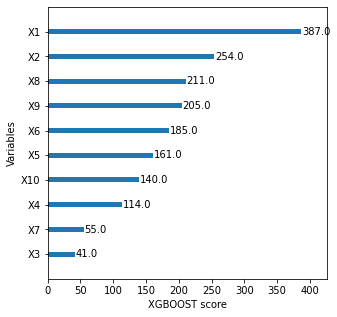}
	\end{subfigure}
		\begin{subfigure}[b]{0.4\textwidth}
		\centering
		\includegraphics[scale=0.6]{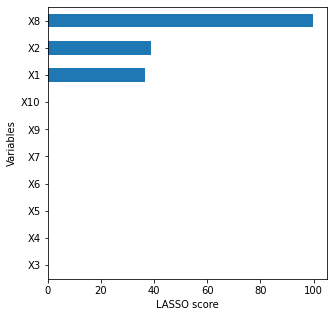}
	\end{subfigure}
	
		\begin{subfigure}[b]{0.4\textwidth}
		\centering
		\includegraphics[scale=0.6]{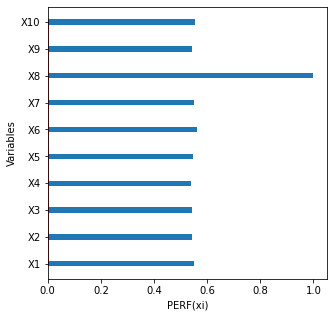}
	\end{subfigure}
		\caption{Variable score on data with highly correlated features }
		\label{fig 4.2}

	\begin{table}[H]
	    \centering
	    \scalebox{0.8}{\begin{tabular}{|rrrrrrrrrrr|}
	        \toprule &  XGBOOST feat. &  PERF feats. &    SVM feats. &  LASSO feats. &     RF feats. &  All features\\
	        \midrule Mean square error       &   2.044925e-27 &  8.791317e-28 &  1.917775e-27 &  1.917775e-27 &  9.027497e-28 &  3.197355e-27 \\
	        Mean absolute error     &   3.414104e-14 &    2.542074e-14 &  3.427561e-14 &  3.427561e-14 &  2.192186e-14 &  4.431472e-14 \\
	        Root Mean Squared Error &   4.522084e-14 &   2.965016e-14 &  4.379240e-14 &  4.379240e-14 &  3.004579e-14 &  5.654516e-14 \\
	   \bottomrule
	    \end{tabular}}
	    	\caption{Errors of each VIMs using its most important features on the high correlated data}
	    \label{table 4.2}
	\end{table}
\end{figure}


\begin{figure}[H]

	\centering
	
	\begin{subfigure}[b]{0.4\textwidth}
	\centering
    \includegraphics[scale=0.55]{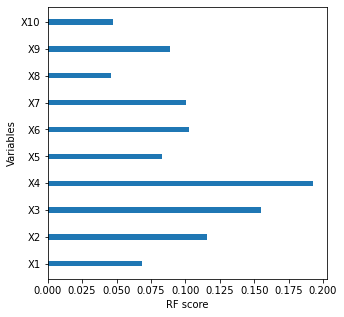}
	\end{subfigure}
	\begin{subfigure}[b]{0.4\textwidth}
		\centering
		\includegraphics[scale=0.55]{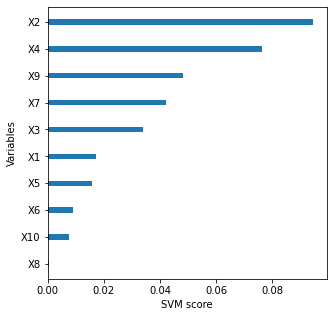}
	\end{subfigure}

	\centering
	\begin{subfigure}[b]{0.4\textwidth}
		\centering
		\includegraphics[scale=0.55]{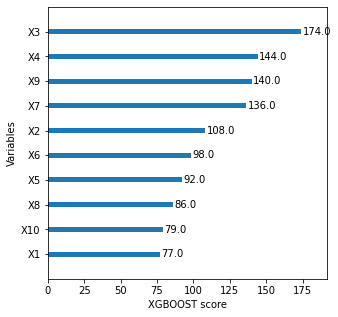}
	\end{subfigure}
		\begin{subfigure}[b]{0.4\textwidth}
		\centering
		\includegraphics[scale=0.55]{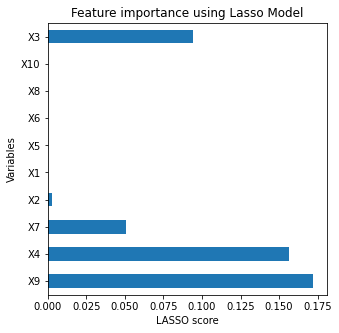}
	\end{subfigure}
	
		\begin{subfigure}[b]{0.4\textwidth}
		\centering
		\includegraphics[scale=0.55]{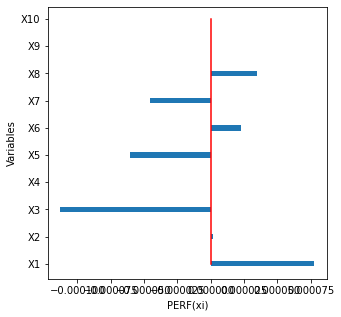}
	\end{subfigure}
	\caption{Variable score on data with redundant features }
	\label{fig 4.3}
	
\begin{table}[H]
			\scalebox{1}{\begin{tabular}{|cccccccc|}
					\toprule &  XGBOOST feat. &  PERF feats. &  SVM feats. &  LASSO feats. &  RF feats. &  All features \\
					\midrule
					Accuracy score &       0.793939 &     0.690909 &    0.800000 &      0.793939 &   0.800000 &      0.793939 \\
					Recall         &       0.722543 &     0.624277 &    0.716763 &      0.722543 &   0.716763 &      0.728324 \\
					Precision      &       0.862069 &     0.744828 &    0.879433 &      0.862069 &   0.879433 &      0.857143  \\
					\bottomrule
			\end{tabular}}
			\caption{Errors of each VIMs using its most important features on the high redundant data}
			\label{table}
		\end{table}	
	
	\end{figure}

\begin{figure}[H]

	\centering
	
	\begin{subfigure}[b]{0.4\textwidth}
	\centering
    \includegraphics[scale=0.55]{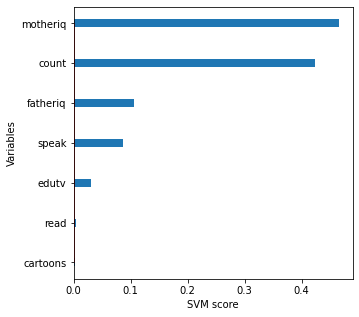}
	\end{subfigure}
	\begin{subfigure}[b]{0.4\textwidth}
		\centering
		\includegraphics[scale=0.55]{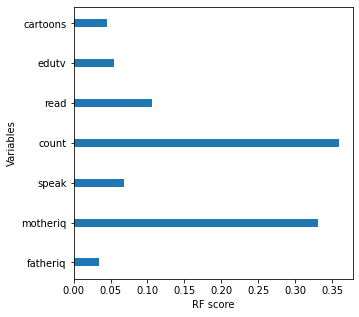}
	\end{subfigure}

	\centering
	\begin{subfigure}[b]{0.4\textwidth}
		\centering
		\includegraphics[scale=0.55]{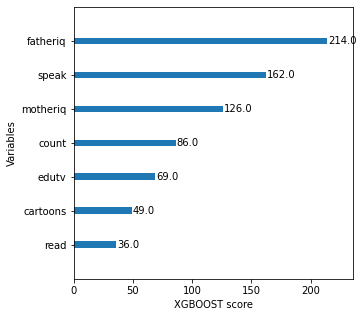}
	\end{subfigure}
		\begin{subfigure}[b]{0.4\textwidth}
		\centering
		\includegraphics[scale=0.55]{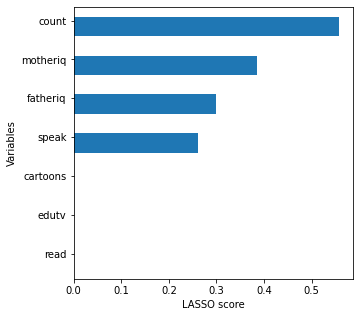}
	\end{subfigure}
	
		\begin{subfigure}[b]{0.4\textwidth}
		\centering
		\includegraphics[scale=0.55]{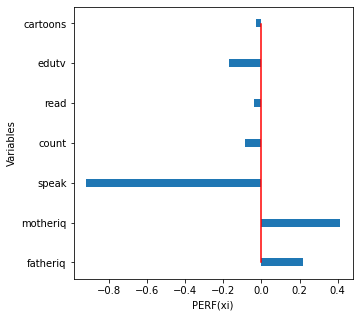}
	\end{subfigure}
	\caption{Variable scores on the gifted data set}
	\label{fig 4.4}
	
		\begin{table}[H]
			\scalebox{1}{\begin{tabular}{|ccccccccc|}
					\toprule
					 &  XGBOOST feat. &  PERF feats. & SVM feats. &  LASSO feats. &  RF feats. &  All features& \\
					\midrule
					Mean square error 	 & 19.503736 &   9.252851 & 5.935797&     5.935797& 5.935797& 6.125717\\ 
					Mean absolute error &  3.607770 &  2.558122 &  1.954763 	 &1.954763 	 & 1.954763 	&   1.976887 \\
					Root Mean Squared Error & 4.416303 	 &  3.041850 &  2.436349 &    2.436349 	 &  2.436349 &  2.475019 \\
					\bottomrule
			\end{tabular}}
			\caption{Errors of each VIMs using its most important features on the gifted data set}
			\label{table 4.1}
		\end{table}
	
	\end{figure}

\begin{figure}[H]

	\centering
	
	\begin{subfigure}[b]{0.4\textwidth}
	\centering
    \includegraphics[scale=0.6]{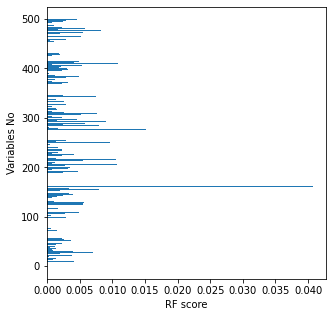}
	\end{subfigure}
	\begin{subfigure}[b]{0.4\textwidth}
		\centering
		\includegraphics[scale=0.6]{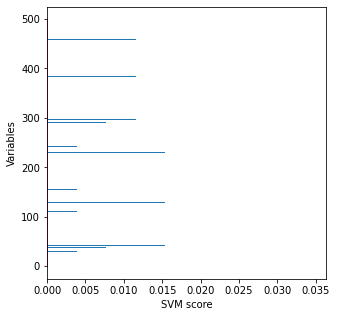}
	\end{subfigure}

	\centering
	\begin{subfigure}[b]{0.4\textwidth}
		\centering

		\includegraphics[scale=0.6]{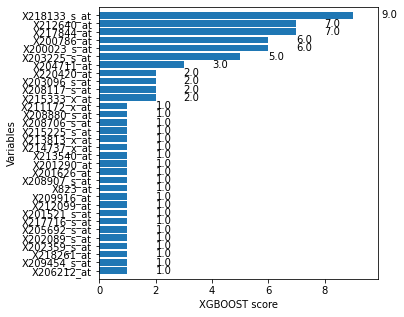}
	\end{subfigure}
		\begin{subfigure}[b]{0.4\textwidth}
		\centering
		\includegraphics[scale=0.6]{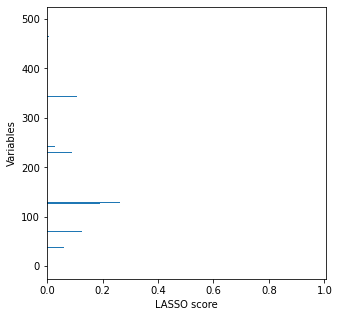}
	\end{subfigure}
	
		\begin{subfigure}[b]{0.4\textwidth}
		\centering
		\includegraphics[scale=0.6]{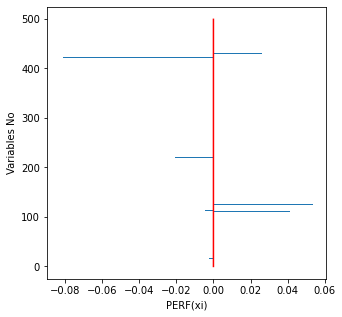}
	\end{subfigure}
		\caption{Variable score on the prostate-cancer dataset }
		\label{fig 4.5}

	\begin{table}[H]
	\centering
			\scalebox{1}{\begin{tabular}{|c|c|c|c|c|c|}
					\toprule
					Methods &  XGBOOST VIMs &  PERF VIMs & SVM VIMs  &  LASSO VIMs  &  RF VIMs \\
					\midrule
					Running time in second	 & 0.72578 &   0.77333 & 11.07807&     2.80276& 0.89753\\ 
					\bottomrule
			\end{tabular}}
			\caption{Computational time of each methods on the prostate-cancer dataset}
			\label{table 4.4}
		\end{table}	
\end{figure}

\section{Discussion and conclusion}
\label{sec3:headings}
\subsection{Discussion of the results}

All the five methods of VIMs estimation used in this essay are sometimes efficient and deficient depending on the characteristic of the datasets. PERF and XGBOOST VIMs are data demanding ie they do not really produce good results when the sample size is small while SVM, LASSO, and RF do quite well and generally agree on the choice of the most important features. This is illustrated in Table \ref{table 4.1} for the gifted dataset.
For highly correlated data, PERF and RF are very suitable to identify important features. According to the simulation  we did, PERF reduces the MSE of our neutral model by more than ten times and was followed closely by RF. The other methods are doing a great job as well in this situation and all the most important variables chosen by each of the methods reduce the MSE error of the neutral model. Table \ref{table 4.2} gives the overview of the score of the different methods applied to the simulated data with highly correlated features.
Concerning data with many redundant features, RF and SVM are the most appropriate in this case. In the context of our simulation, they increase the precision of our neutral model from 0.7 to 0.8. Also, XGBOOST and LASSO are suitable when there is redundant features in the data while the PERF does not provide good results in this case for the data with redundant features we simulated. The variables chosen by PERF reduced the accuracy of the neutral model by more than ten percent as can be seen in Table \ref{table}.  
When it comes to dealing with ultra-high data, all the methods did well and among the 500 features used in the prostate cancer datasets, all the VIMs techniques retain at most 
seven features but in terms of the computational speed, XGBOOST is the fastest closely followed by PERF and RF as shown in Table \ref{table 4.4} while SVM has the slowest computational time.
In terms of interpretability, PERF is the best because of its natural straight red line at point 0. That line makes the selection and interpretation of feature scores very easy. With that straight line, useful features are those on the right side of the line and all others features on the left or on the line are not important. This makes things simple contrary to the four other methods where we need to define a threshold to choose appropriate features.
The strength and the weakness of the five methods have been summarized in the table below:
\newpage
\begin{figure}[H]
    \centering
    \includegraphics[scale = 0.7]{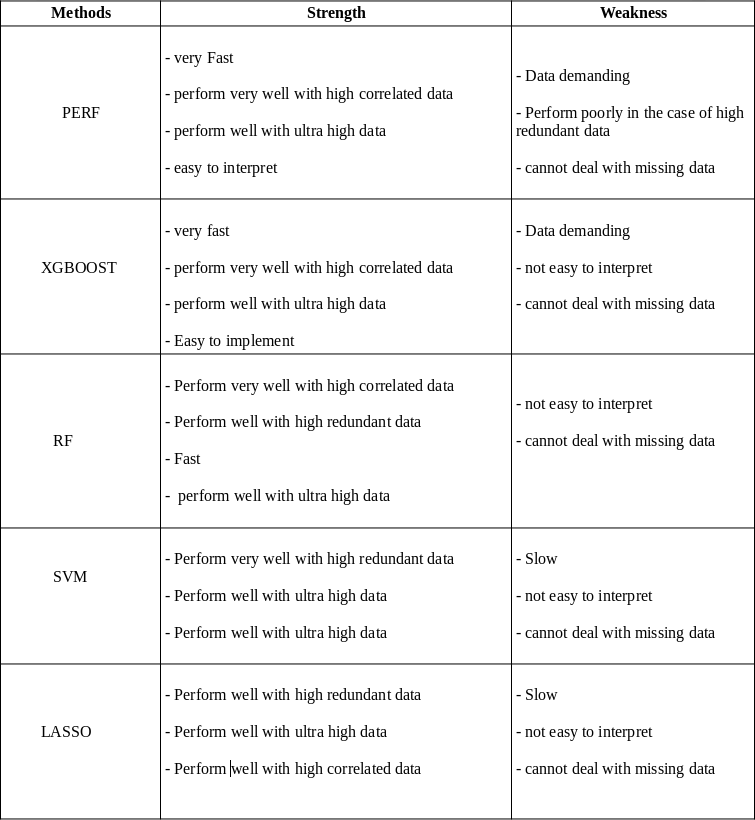}
    \caption{Strength and weakness of each of the methods}
    \label{fig:my_label}
\end{figure}
\subsection{Conclusion}
We have discovered in the framework of this thesis that concerning modern methods of estimating the importance of variables, there are no perfect methods. We have evaluated the performance of five powerful methods of VIMs, four of which are widely used in machine learning and one recently developed by \cite{fokoue2015prediction}, using different types of data: High correlated, Ultra-high, High redundant, and infra high. We have found that although PERF is not recommended when the sample size is small but it is the most suitable in the case of highly correlated data as well as RF. SVM is the best when it comes to high redundant data but it had the slowest execution time, XGBOOST has the fastest execution time and as PERF, it is not safe to use it when the sample size of the data is small. LASSO and RF are very versatile, although they are not giving the best result, they are suitable in almost all situations. All the five methods we have studied faced a common issue which is that they are not able to deal with data containing missing values it is in this way that \cite{hapfelmeier2014new} have defined a new variable importance measure for the random forest with missing data which is able to handle data with or without missing value.

\newpage
\bibliographystyle{unsrt}  
\bibliography{references}

\end{document}